# Integration of neural network and fuzzy logic decision making compared with bilayered neural network in the simulation of daily dew point temperature

Guodao Zhang[a], Shahab S. Band[b], Sina Ardabili[c], Kwok-Wing Chau[d], Amir Mosavi[e,f]

[a] College of Computer Science and Technology, Zhejiang University of Technology, Hangzhou 310023, China;
[b] National Yunlin University of Science and Technology, Yunlin 64002, Taiwan;
[c] Department of Informatics, J. Selye University, Komarom, Slovakia;
[d] Department of Civil and Environmental Engineering, The Hong Kong Polytechnic University, Hung Hom, Kowloon, Hong Kong, China;
[e] John von Neumann Faculty of Informatics, Obuda University, Budapest, Hungary;
[f] Slovak University of Technology in Bratislava, Slovakia

**Abstract**

In this research, dew point temperature (DPT) is simulated using the data-driven approach. Adaptive Neuro-Fuzzy Inference System (ANFIS) is utilized as a data-driven technique to forecast this parameter at Tabriz in East Azerbaijan. Various input patterns, namely T min, T max, and T mean, are utilized for training the architecture whilst DPT is the model's output. The findings indicate that, in general, ANFIS method is capable of identifying data patterns with a high degree of accuracy. However, the approach demonstrates that processing time and computer resources may substantially increase by adding additional functions. Based on the results, the number of iterations and computing resources might change dramatically if new functionalities are included. As a result, tuning parameters have to be optimized inside the method framework. The findings demonstrate a high agreement between results by the data-driven technique (machine learning method) and the observed data. Using this prediction toolkit, DPT can be adequately forecasted solely based on the temperature distribution of Tabriz. This kind of modeling is extremely promising for predicting DPT at various sites. Besides, this study thoroughly compares the Bilayered Neural Network (BNN) and ANFIS models on various scales. Whilst the ANFIS model is extremely stable for almost all numbers of membership functions, the BNN model is highly sensitive to this scale factor to predict DPT.

**Introduction**

Dew point temperature (DPT) is determined by the temperature at which air turns into liquid water owing to a high amount of water molecules (Ali, Fowler, & Mishra, 2018). A precise DPT estimation is essential in agricultural activities, such as in determining the quantity of accessible moisture in air and predicting near-surface humidity. DPT and humidity levels are frequently employed to determine the amount of humidity in the air (Famiglietti, Fisher, Halverson, & Borbas, 2018). DPT could also be used to determine crop temperature in the presence of glaciation. Several researches have been conducted to determine this parameter and provide modeling of this parameter. Their models were performed using a variety of inputs, and this parameter was predicted based on experimental results (Shiri, 2019). There are many numerical and analytical approaches to estimate these factors in various areas, and many investigations have explored these analyses for DPT prediction (Baghban, Bahadori, Rozyn, Lee, Abbas et al., 2016; Shank, McClendon, Paz, & Hoogenboom, 2008; Zounemat-Kermani, 2012). However, owing to the availability of datasets for DPT, machine learning techniques are popular among all numerical methods and analyses. Because machine learning algorithms are capable of collecting historical data, there is a strong possibility of developing data-driven approaches based on this kind of modeling.

Artificial Neural Network (ANN), Adaptive neuro-fuzzy inference system (ANFIS), Gene Expression Programming (GEP), and other machine learning techniques, for example, are used to train datasets and subsequently forecast DPT. ANFIS is an acceptable toolbox among all techniques for predicting these parameters. ANFIS integrates intelligent characteristics into an ensemble learning framework and merges these capabilities into a supervisory control system that is able to adjust spontaneously (Babanezhad, Nakhjiri, & Shirazian, 2020; Razavi, Sabaghmoghadam, Bemani, Baghban, Chau et al., 2019; Sefeedpari, Rafiee, Akram, Chau, & Komleh, 2015). The primary concept behind the soft computing strategy is to accumulate input-output pairs that include both values, then train the recommended network using these input-output pairs. ANFIS method is a practical methodology for structuring fuzzy inference systems when they have input/output data with a couple of information given. Fuzzy logic, combined with a feature called fuzzy tuning, allow one to alter membership function settings so as to enable the best connected fuzzy inference ability to track the given input/output data. The approach for forecasting process parameters is also referred to when ANFIS model is taken into account. The neural and fuzzy logic networks assist this application to learn how it works and then predict how it works in the future. The tuning options of this algorithm enable it to be a flexible toolbox for high level prediction of the process. The process for designing this model may begin with various combinations of input, functions, and rules in order to develop the whole structure of the model (Jang, 1993; M. Pourtousi, Zeinali, Ganesan, & Sahu, 2015; Razavi, Sabaghmoghadam, Bemani, Baghban, Chau et al., 2019; Yan, Safdari, & Kim, 2020). ANFIS has been successfully employed by several researchers for different applications. Citakoglu et al (2014) employed ANFIS for the prediction of monthly mean reference evapotranspiration in the presence of long-term average monthly climatic data. Cobaner et al (2014) employed ANFIS in comparison with ANN for predicting the mean monthly air temperature in the presence of metrological data in Turkey. Results have been evaluated using error values. Citakoglu (2015) in a study, employed ANFIS technique for predicting the solar radiation and compared that with ANN and multiple linear regression (MLR) in the presence of meteorological data at monthly periods. Citakoglu (2017) employed ANFIS technique in comparison with ANN for predicting the soil temperatures at 100-cm depths below the soil surface. In order to examine DPT in Tabriz, three distinct input factors are taken into consideration. These parameters are utilized to construct a machine learning algorithm for the prediction of DPT based on the input parameters. For training the data, about 65% of the whole data is used. For the validation phase, the remaining 35% of data is accounted. To ensure the comparison of the actual data and machine learning results are being made at each level of research, simulations are used for every level of research. Before the training phase begins, random samples of the data are utilized for training the model. This randomization of the parameters is used to assure that no association exists between the input parameters. In the testing step, 35% of the training datasets are picked. The decoupling of testing and training occurs at this stage. Reducing processing time and irrelevant input parameters are also studied in this investigation. Besides, the degree of membership function in each input parameter is shown to examine the effect of function characteristics in each input parameter. This study undertakes to examine the effect of membership functions in each input parameter, as well as the impact of inputs on the output of this model. For further comparison, the ANFIS approach is compared to the Bilayered Neural Network (BNN) model for DPT prediction. The BNN model takes into account a different number of layers, whereas the ANFIS model takes into account a different number of membership functions in input parameters.

**Materials and Methods**

Figure 1(a) shows the summary of the study. The figure indicates that the membership functions and their characteristics, including their number and type, are selected for each input in the very beginning. Then the data is selected, and how much data needs to be engaged in testing and training are determined in the next stage. When the system reaches a high prediction capability, the prediction process starts, and the predicted data is compared to the data which is not applied in the study (not seen by AI model). As shown in the figure, if each model has a high level of error, the membership function and the data are changed to achieve higher accuracy in prediction and low number of error. This stage is repeated in a trial and error manner to achieve a higher prediction capability, and therefore, the model is selected.

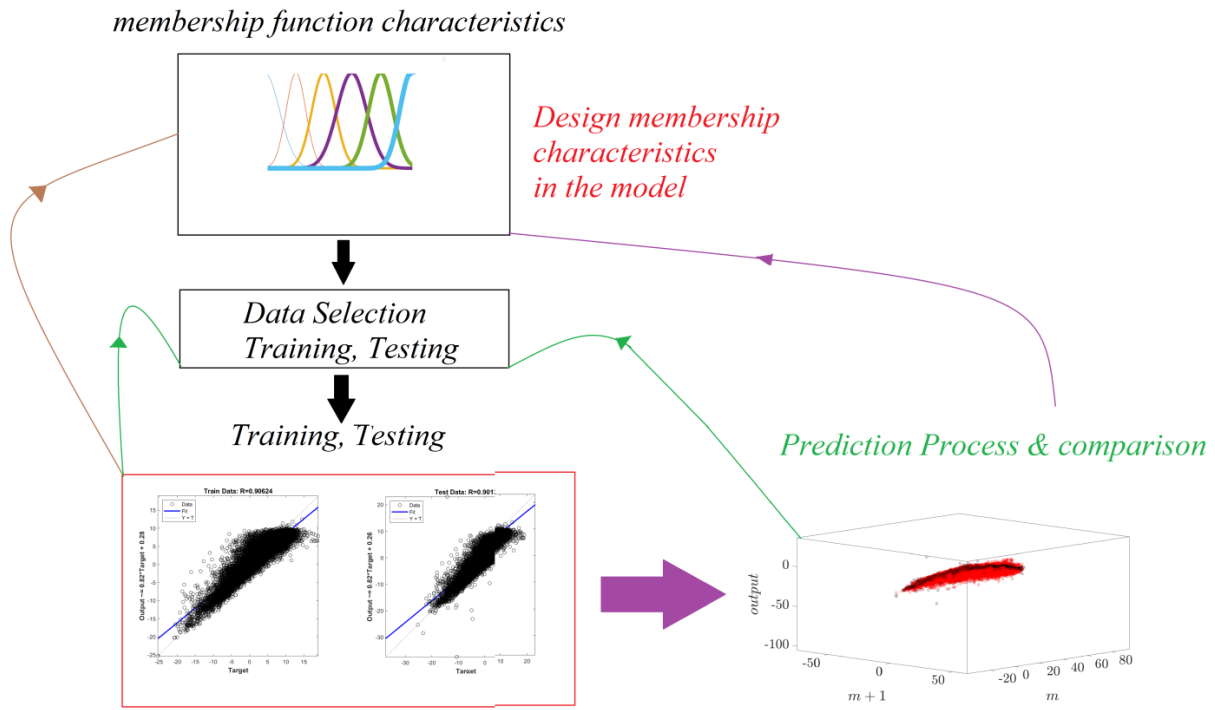

Figure 1(a): Schematic summary of the model for training and testing. The data selection process is shown in this schematic figure.

*ANFIS method*

ANFIS, or Adaptive Neuro-Fuzzy Inference System, was proposed by Takagi-Sugeno Fuzzy Inference System. Using this method, the membership functions in the learning phase could be altered, and therefore, the parameters can be changed based on the optimization process (Vaidhehi, 2014). The method is popular with its broad uses in different fields. It was applied in psychological aspects of teaching contexts (Z. Pourtousi, Khalijian, Ghanizadeh, Babanezhad, Nakhjiri et al., 2021), different dimensions of learners (Bachtiar, Sulistyo, Cooper, & Kamei, 2017), prediction of various diseases such as cancer (Kalaiselvi, & Nasira, 2014) or even diabetes (Kirisci, Yılmaz, & Saka, 2019), weather forecasting (Tektaş, 2010), wind speed prediction (Liu, Tian, & Li, 2015), etc. It is also worth mentioning that for predicting the wind speed, other models, such as support vector machines and Gaussian process regression, can be applied (Başakın, Ekmekcioğlu, Çıtakoğlu, & Özger, 2021). However, using ANFIS seems to be an efficient computation approach. Besides, it can be used when complex parameters exist in the system. The first layer of the model consists of input parameters of MFs. Three different inputs are considered in this work, namely T max, T min and T mean, while the output of the model is DPT. The next layer is the membership layer, which receives the signals from the previous layer. In the following equation, μAi, μBi, and μCi are the input parameters. In the ANFIS model, the membership function of each input parameter is defined to translate datasets into the fuzzy structure.

$$w_i = \mu_{Ai}(m)\, \mu_{Bi}(m+1)\mu_{ci}(m+2) \tag{1}$$

The rule layer refers to the third layer, and the nodes denote the normalized weights. Besides, Equation (2) shows the firing strengths $w_i$.

$$\overline{w_i} = \frac{w_i}{\sum(w_i)} \tag{2}$$

The next layer is the defuzzification layer or the fourth layer, which uses the If-Then rule's function for providing a particular outcome. In Equation (3), different parameters, namely pi, qi, ri, and si, refers to parameters of the If-Then rules.

$$\overline{w_i}f_i = \overline{w_i}(p_i x + q_i y + r_i t + S_i) \tag{3}$$

*Bilayered neural network*

In this study, the architecture of multilayer perceptron network (MLP) is used. This network architecture has two layers (Bilayered). The architecture of the MLP network is known as one of the most common types of ANNs that is easily available in many computing software. This type of network is used for supervised forecasting (Ackora-Prah, Sakyi, Ayekple, Gyamfi, & Acquah, 2014). In this type of network, each neuron is considered as a computational unit and creates input at the output by linearly combining the input signals. In this network, the activation function also plays an important role. The learning algorithm in this study is the Levenberg-Marquardt algorithm. This algorithm integrates steepest decent technique and Gauss-Newton algorithm. The advantages of this method include enhanced training rate as well as high stability and reliability (Reynaldi, Lukas, & Margaretha, 2012; Smith, Wu, Wilamowski, & systems, 2018). This algorithm uses the sum of squares of errors in the validation phase.

*Data*

The Tabriz synoptic station, located in East Azerbaijan, is one of the earliest Iranian meteorological stations. All DTP datasets for various years are collected from this location (1998-2016). The location of Tabriz in Iran and further information regarding the station and datasets can be found in previous studies (Qasem, Samadianfard, Sadri Nahand, Mosavi, Shamshirband et al., 2019).

When examining a model developed utilizing ANFIS methodology, one may describe it as a learning algorithm for forecasting dew point temperature on a daily basis. This program can learn the process and then predict the daily dew point temperature using the neural network capability and the judgment section of fuzzy logic. ANFIS is particularly effective for anticipating the behavior of nonlinear and complex systems. In each instance of simulation, a different number of iterations is utilized in order to achieve numerical convergence, and the gaussmf function is employed for each input parameter. A more in-depth review of ANFIS approaches can be found in different prior publications (Famiglietti, Fisher, Halverson, & Borbas, 2018; Shiri, 2019). Four and six membership functions are respectively evaluated in each input to test the prediction capability. For the purpose of comparison, the ANFIS approach is compared to the Bilayered Neural Network (BNN) model for DPT prediction. The BNN model considers different number of layers, whereas the ANFIS model considers different numbers of membership functions in input parameters.

**Results**

Sixty-five percent of data is used for training, and the remaining 35% of data is used for testing and validation. At each level of investigation, a comparison is made between the actual data and machine learning results. Before the training phase starts, the data used to train the model is randomly selected. To guarantee that no relationship exists between the input parameters, randomization of the parameters is undertaken. 35% of the training datasets are selected for testing in the testing stage. During this stage, testing and training processes are totally decoupled. A number of other elements are taken into consideration. To investigate DPT in Tabriz, three distinct input parameters are analyzed. These input parameters are utilized to construct a DPT prediction machine learning technique. T min, T max, and Tm are the inputs 1, 2, and 3, respectively. T min and T max are renamed r and z, respectively, while DPT is shown as the C factor. Different membership functions are studied in the ANFIS model in order to improve the model's model development. Each input parameter is then translated into a fuzzy structure. It is found that, when using six input parameters, the model's performance is the best. Figure 1(b) illustrates the domain of each input parameter covered by four membership functions. To find out which function each input follows, each input is examined to see which membership functions it takes on (Figures 2 and 3). T max functions may be shown to be evenly distributed over the domain. When applied to T min functions, the variance is non-uniform. This is an example of how the distribution of T max shows the influence of the input on the output.

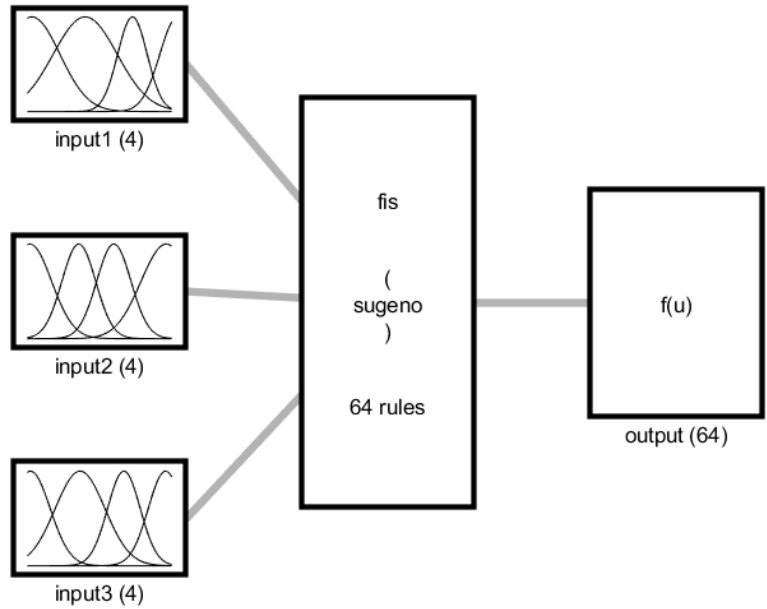

Figure 1(b): ANFIS structure with four membership functions for each input parameter.

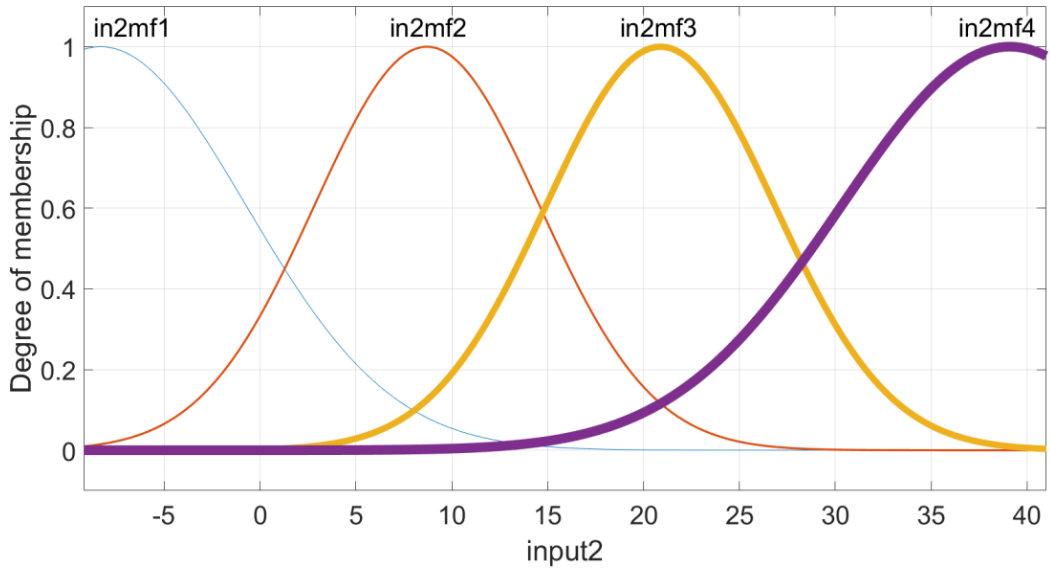

Figure 2: Degree of membership functions as a function of Tmax in Tabriz (with four membership functions).

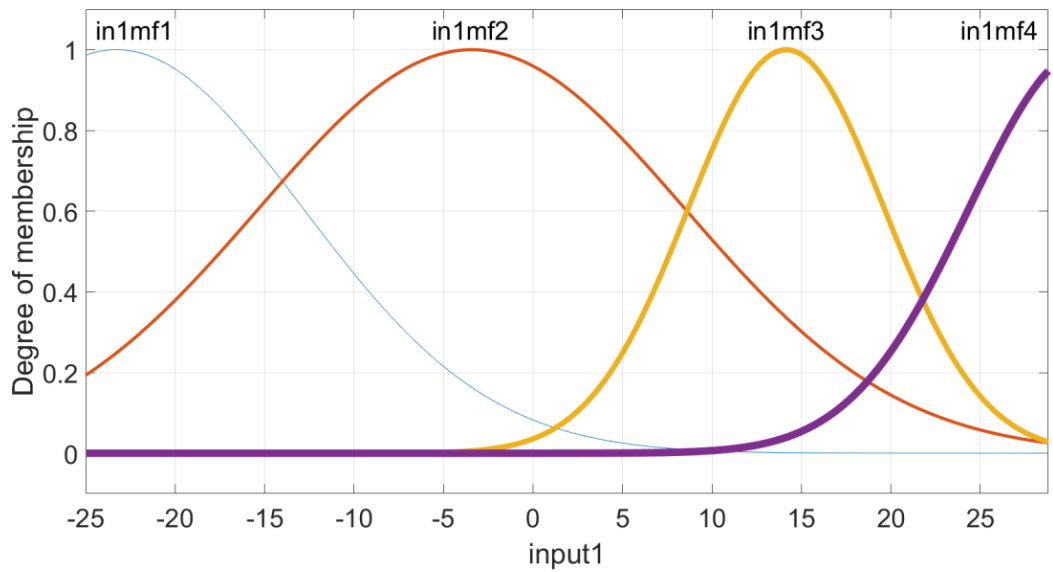

Figure 3: Degree of membership functions as a function of Tmin in Tabriz(with four membership functions).

65% of the entire dataset is used to train the model, and then it is tested against the real data in Tabriz. There is a solid degree of accuracy and prediction capacity (R>0.9) in the outcomes of training (See Figure 4). However, the model's predictive potential is shown throughout the testing procedure (see Figure 5). Datasets (more than 15,000 samples) are compared to their experimental values to provide a clearer picture (See Figure 6). According to the findings, ANFIS has the ability to detect patterns in data with near-perfect accuracy.

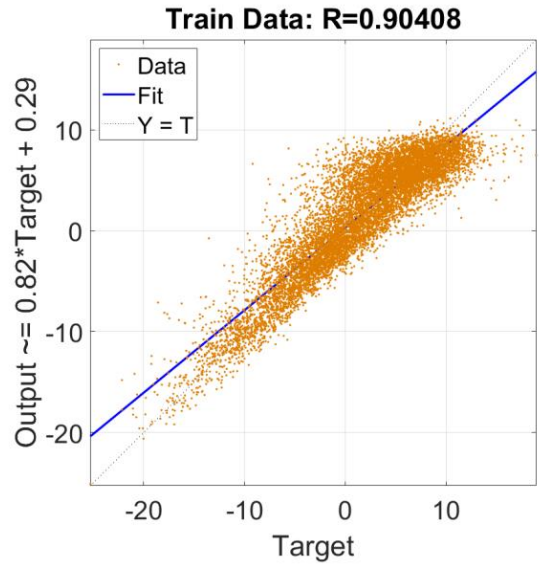

Figure 4: Evaluation of training process with four membership functions in each input.

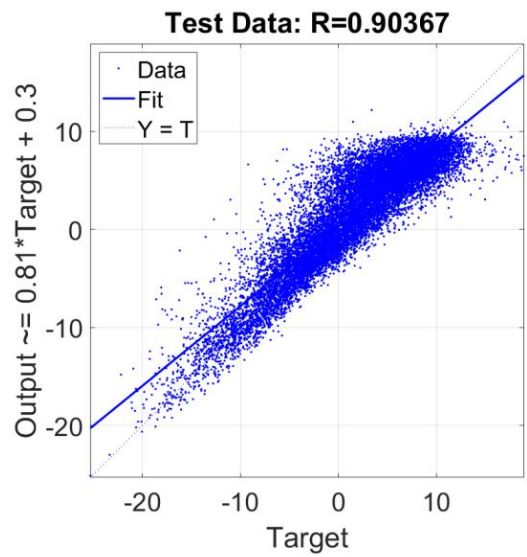

Figure 5: Evaluation of testing datasets with four membership functions in each input.

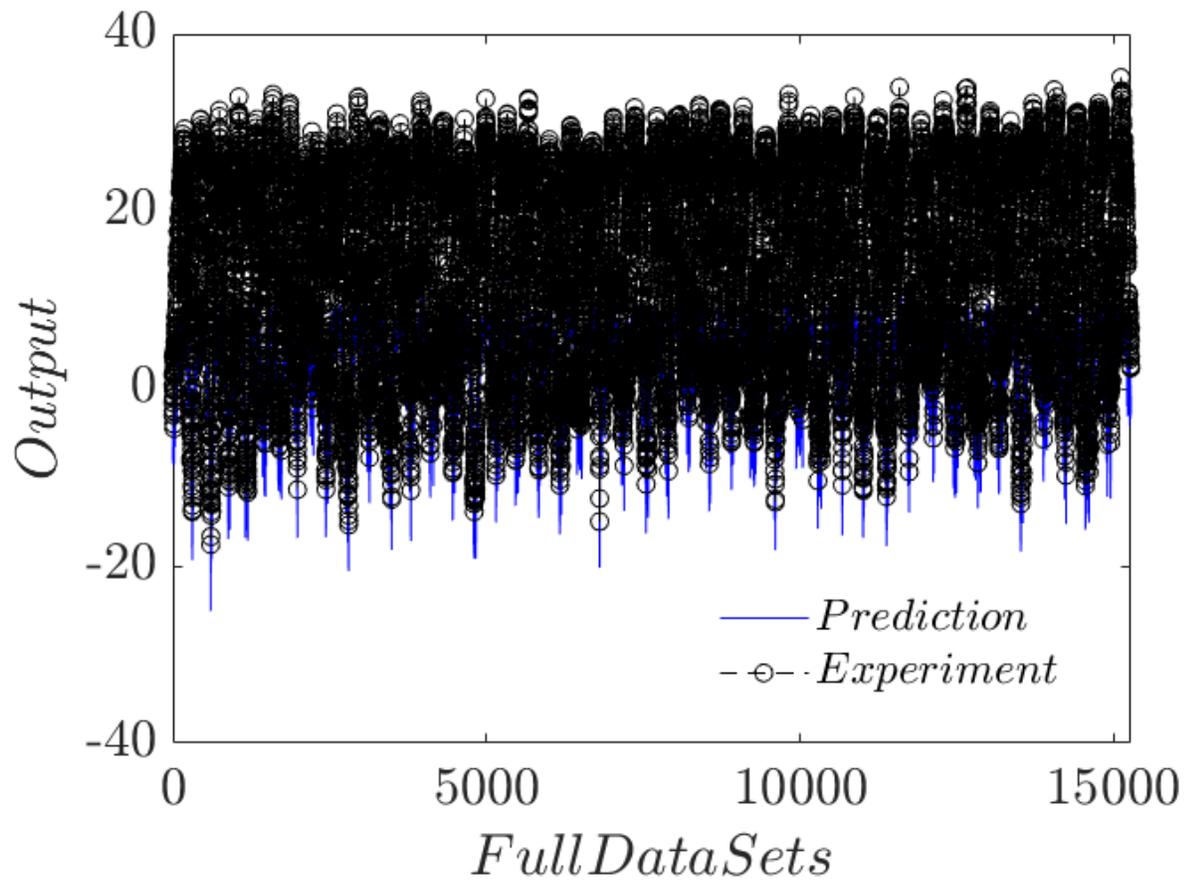

Figure 6: Evaluation of prediction mode with four membership functions in each input.

Figure 7 depicts the domain of each input parameter covered by six membership functions. The function that each input follows is determined by examining each input to see which membership functions it belongs to (Figures 8 and 9). It is possible to demonstrate that T max functions are uniformly distributed over the domain. When applied to T min functions, the variance does not follow a uniform distribution. As an example, consider the following: the distribution of T max demonstrates the effect of an input on the output. The distribution of these types of functions is almost comparable to the distribution of four membership functions. But when compared with that of four membership functions, the non-uniform distribution of the membership function in input is clearer.

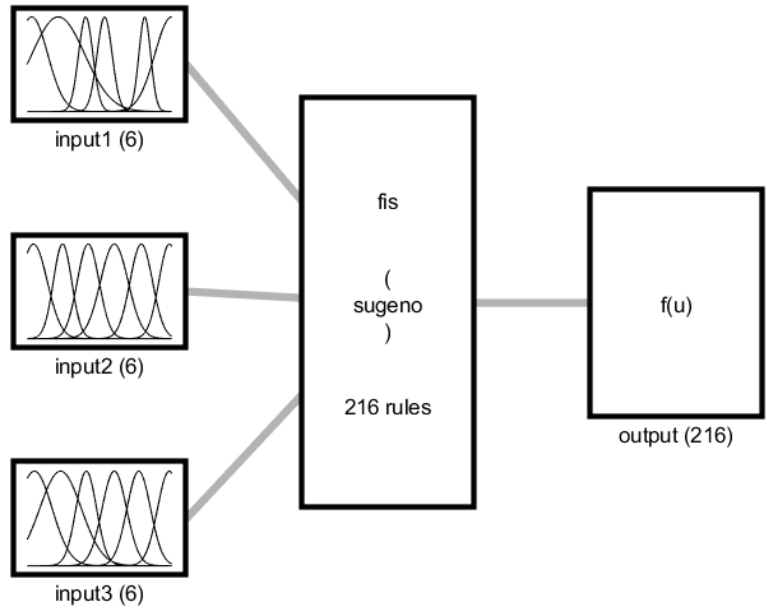

Figure 7: ANFIS structure with membership functions for each input parameter.

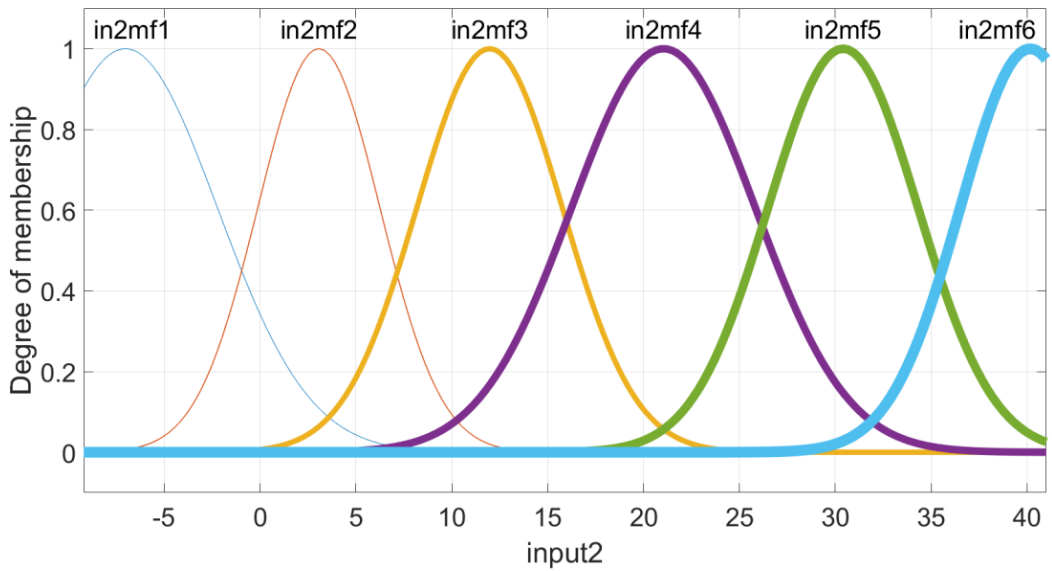

Figure 8: Degree of membership functions as a function of Tmax in Tabriz with membership functions

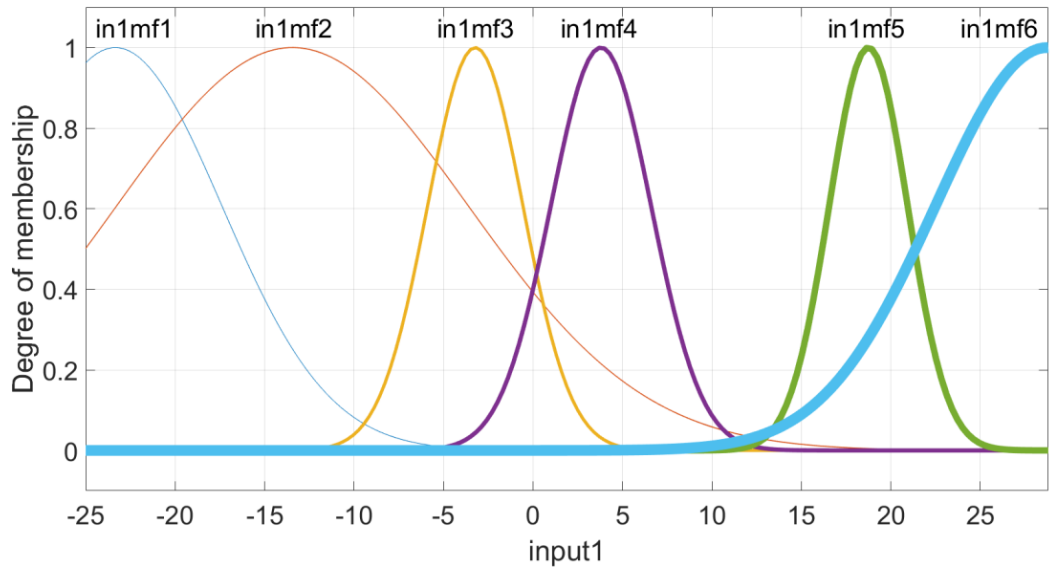

Figure 9: Degree of membership functions as a function of Tmin in Tabriz with membership functions

Using 65 percent of the entire datasets for the training stage, this model with six membership functions in each input parameter is assessed, and then the model is compared with observed values in Tabriz for comparison. Following the training, the participants demonstrate a high degree of agreement and predictability (R>0.9) among themselves and with one another (See Figure 10). Despite that, this model exhibits an extremely high degree of predictability during the whole testing process (see Figure 11). In order to give a more realistic comparison, all datasets are compared with experimental values (See Figure 12). The results suggest that the ANFIS methodology is capable of generally identifying the pattern of data with a high degree of accuracy and that this capability is shown in the experiments. This result is remarkably close to the one obtained with four membership functions in each input parameter. However, it is possible to dramatically increase the computational time and computer resources by including additional functions.

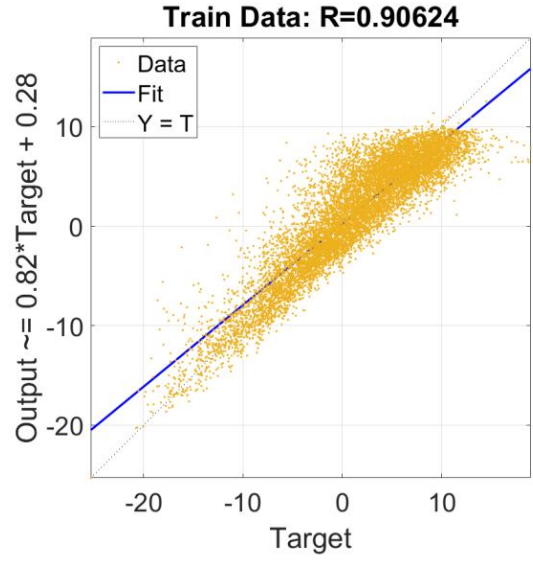

Figure 10: Evaluation of training process with membership functions in each input.

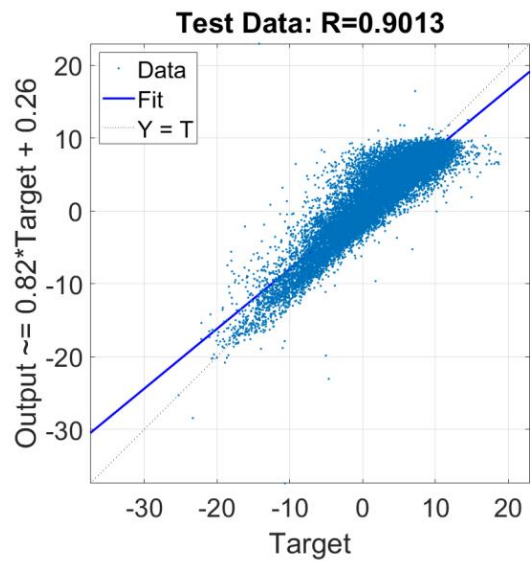

Figure 11: Evaluation of testing process with membership functions in each input.

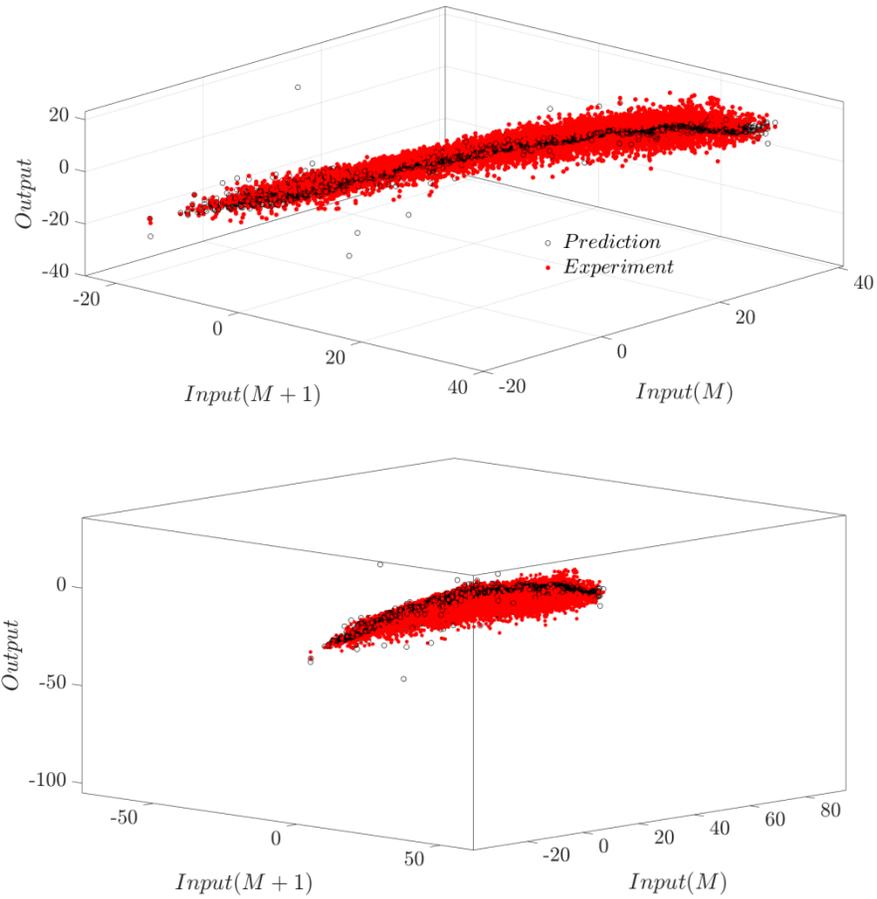

Figure 12: Evaluation of prediction mode with membership functions in each input & comparison with observed data.

The ANFIS model is also tested for different numbers of numerical iterations. The findings indicate that for a short number of iterations, both training and testing processes have high RSME, but as the number of iterations increases, RSME decreases. The number of numerical iterations in the testing procedure for both MMSE and RSME rises between 50 and 100. These parameters, however, diminish after around 100 iterations (See Figure 13).

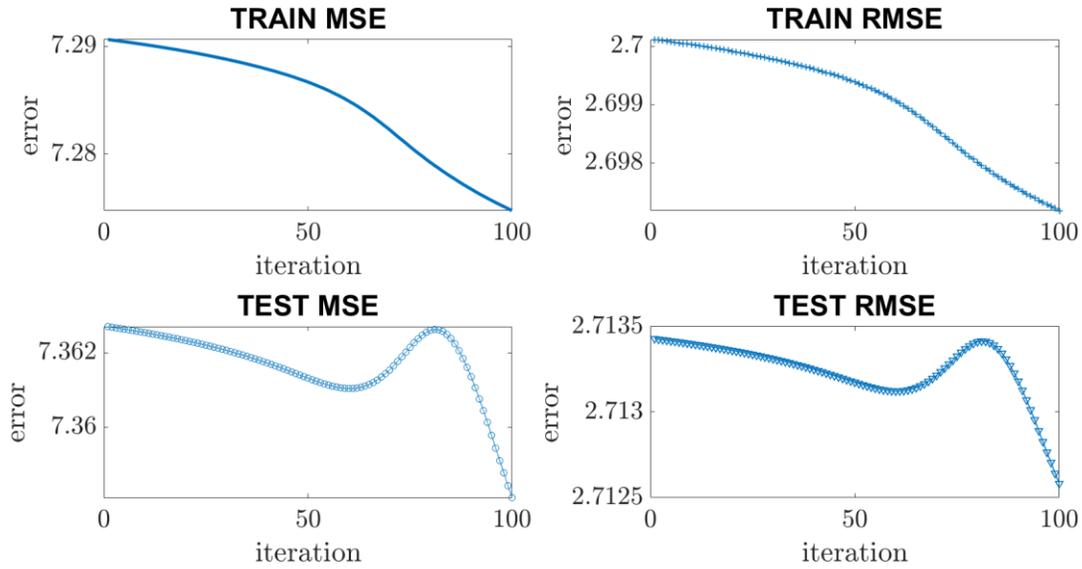

Figure 13: Assessment of AI model for different number of iterations. Training and testing are evaluated against experimental findings based on RSME and MSE.

In this research, the proportion of training datasets is investigated for further comparison. Table 1 depicts the effect of various dataset sizes on the training process. The findings indicate that increasing this parameter has no effect on model correctness, suggesting that increasing the number of datasets in the training process is pointless.

Table 1: Different amounts of datasets during training process and evaluation of model, based on RSME and MSE criteria.

| Data Selection | Train MSE | Train RMSE | Test MSE | Test RMSE | R-Train | R Test |
|---|---|---|---|---|---|---|
| 70% | 7.23 | 2.68 | 7.34 | 2.71 | 0.9 | 0.9 |
| 75% | 7.28 | 2.69 | 7.37 | 2.71 | 0.9 | 0.9 |
| 85% | 7.17 | 2.67 | 7.31 | 2.7 | 0.9 | 0.9 |
| 95% | 7.31 | 2.70 | 7.30 | 2.7 | 0.9 | 0.9 |

Apart from the ANFIS model, a Bilayered Neural Network (BNN) model is utilized to forecast DPT based on three distinct input factors ( Tmin, T max, and T mean). Figure 14 demonstrates that there is a high level of agreement between BNN and experimental observation. However, for certain temperature datasets, the BNN model is extremely near to the average DPT value.

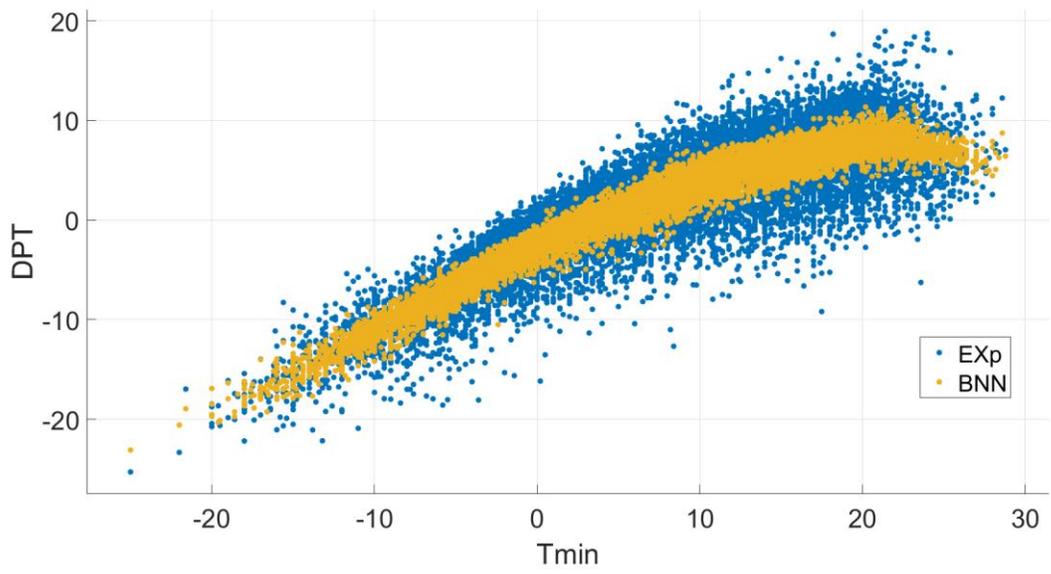
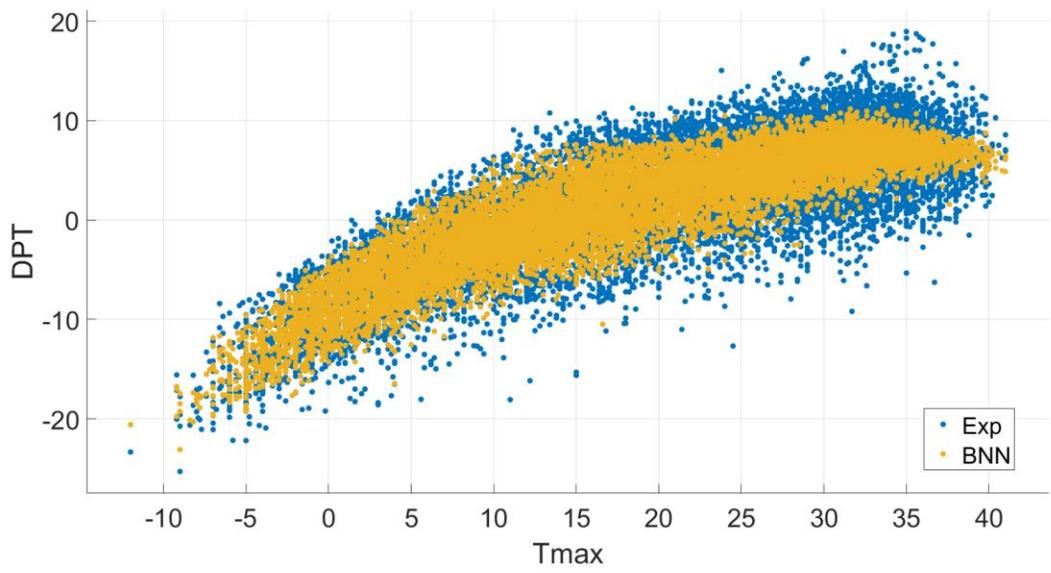

Figure 14: Comparison between BNN and experimental data, when DPT is a function of T min and T max.

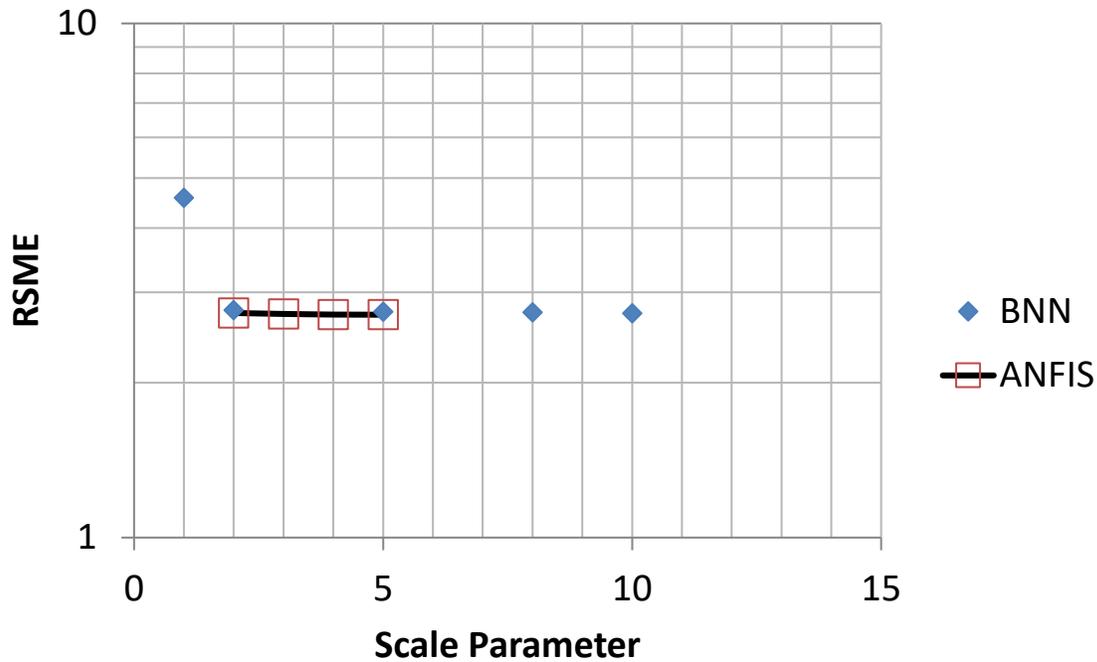

Figure 15: Comparison between BNN and ANFIS method with RSME cretria.

In the last section of this research, BNN and ANFIS models are compared by using RSME criteria at different scale factors. In the BNN model, the scale factor is the number of layers, while in the ANFIS approach, the scale factor is the number of membership functions in each input parameter. The findings indicate that the ANFIS model is very stable for almost all numbers of membership functions. However, the BNN model is very sensitive to this scale factor in order to achieve a high degree of accuracy (See Figure 15).

**Conclusions**

In this study, data-driven simulation is used to model dew point temperature (DPT). The forecasting method based on ANFIS is used to estimate this factor at Tabriz. Additionally, the model's architecture is trained using input patterns, namely T min, T max, and T mean, and DPT is the model's output. Data patterns are found to be accurately identified using the ANFIS technique. The results reveal that the number of iterations and computing resources used might change dramatically if new functionalities are included. When this happens, the tuning settings within the method framework have to be improved. These results show that the machine learning approach (based on machine learning data) correlates with the actual values. This prediction toolbox enables us to compute DPT level at Tabriz solely based on the temperature distribution. This simulation has tremendous potential for forecasting DPT concentrations at different locations. Machine learning observations occur when different input and output parameters are selected, and then machine learning processing begins to explore the connections between these input and output parameters.

ANFIS approach is utilized as a machine learning method to determine how consistent the daily dew point temperature is. This model is also used to examine the interaction connection between input and output components. The results reveal that the ANFIS method is capable of identifying data patterns with a high degree of accuracy in general. This is quite close to the result obtained by employing four membership functions in each of the input parameters. Throughout the course of training and testing, all prediction results show a high degree of prediction (R>0.9). The last section of this research compares the BNN and ANFIS models by employing RSME criteria at different scale factors. In the BNN model, the scale factor is the number of layers, while in the ANFIS technique, the scale factor is the number of membership functions in each input parameter. The findings suggest that the ANFIS model is very stable for almost all numbers of membership functions. The BNN model, on the other hand, is very responsive to this scale factor in order to achieve a high degree of accuracy. It is also worth mentioning that ANFIS has various benefits for researchers. For instance, it is efficient regarding the computations. Moreover, it can be applied when different and complex parameters are engaged in the processes.